\documentclass{article}
\usepackage[preprint]{neurips_2026}
\usepackage[utf8]{inputenc}
\usepackage[T1]{fontenc}
\usepackage{amsmath,amssymb}
\usepackage{graphicx}
\usepackage{booktabs}
\usepackage[hidelinks]{hyperref}
\usepackage{xcolor}
\usepackage{enumitem}
\usepackage{microtype}
\usepackage{tcolorbox}
\usepackage{natbib}

\title{The Growing Pains of Frontier Models:\\When Leaderboards Stop Separating and What to Measure Next}

\author{Adil Amin\\ZEHEN Labs\\{\tt adil@zehenlabs.com}}

\begin{document}
\maketitle

\begin{abstract}
Leaderboards rank frontier models on independent axes but do not reveal
whether capabilities reinforce or trade off across releases---and
at the frontier, this interaction is the more informative signal.
We decompose paired SWE-bench and GPQA Diamond scores into
a population coupling trend and per-release residual ($h$-field) that
diagnoses capability emphasis from two public benchmark scores.
Across 34 models from 10 labs (2024--2026), capabilities cooperate
($r = +0.72$, $p < 10^{-6}$), but cooperation varies systematically:
per-lab coupling slopes span $5\times$ (Google $1.15$ vs.\ DeepSeek $0.23$),
and labs pivot---DeepSeek reversed from reasoning-rich to coding-first
($\Delta h = 15.9$~pp); Anthropic oscillates between coding excursions and recovery.
The population regression serves as an isocline phase boundary:
the same $\sqrt{(a/b)\cdot B_1}$ classifier that identifies the base-scale
coupling transition~\cite{amin2026lying} classifies frontier models
and already detects mixed-phase behavior at the next transition
(two models below the GPQA--IFEval isocline).
The $h$-field is not just diagnostic---it tells you what to change.
Pretraining establishes coupling at $0.871$ while RLHF adds $0.081$~\cite{amin2026lying}:
pretraining-level shifts are permanent (DeepSeek's four-release reversal persists),
post-training shifts are reversible (Anthropic's three coding excursions
each recover within one release),
and inference compute alone shifts $h$ by $+7.8$~pp without retraining.
Knowing which component dominates determines whether to retrain or wait.
We provide a three-step diagnostic (locate, classify, predict),
a per-lab measurement-priority table, and seven falsifiable predictions
with timestamped criteria.
Five post-cutoff releases fall within the 95\% prediction interval.
Code, data, and an interactive dashboard:
\url{https://zehenlabs.com/cape/}.
\end{abstract}

% ════════════════════════════════════════════════════
\section{Introduction}
\label{sec:intro}

When a frontier lab releases a new model, the community asks two questions:
how well does it code, and how well does it reason?
These questions are always asked separately.
SWE-bench measures coding; GPQA Diamond measures reasoning;
each gets its own leaderboard row, its own trajectory, its own narrative.
But no one asks whether improving one helps or hurts the other---and
at the frontier, this interaction turns out to be the more informative signal.

We test it. Using paired benchmark scores (SWE-bench Verified and GPQA Diamond) across
34 frontier models from 10 labs, we measure whether capabilities reinforce or undermine
each other across a lab's release sequence.
Capabilities cooperate ($r = +0.72$, $p < 10^{-6}$),
but the pattern of cooperation varies by lab and over time.
Each lab leaves a fingerprint---a one-number diagnostic that summarizes
whether its models are becoming better reasoners, better coders, or both.

Leaderboards answer ``who is ahead?''
CAPE answers a different question: what kind of progress is this?
A release can climb the leaderboard by moving along the population trend,
by specializing away from it,
or by exhausting an axis that no longer separates frontier models---a
problem now documented across half of all active benchmarks~\cite{akhtar2026plateau}.
Those cases require different responses.
We operationalize this with three quantities available from public scores alone:
population coupling ($r$), release residual ($h$), and saturation ratio ($\sigma$), so
that any practitioner can reproduce the diagnosis from the same model-card numbers.

Where~\cite{amin2026lying} establishes that the coupling transition is engineerable
at base scale, showing that architecture, curation, and distillation
shift the critical threshold, the present work asks:
once models are deep in the cooperative regime,
what determines their trajectory through coupling space,
and what can practitioners do about it?

\noindent\textbf{Contributions.}
\begin{enumerate}[nosep,leftmargin=*]
  \item We measure cooperative frontier coupling on the largest cross-lab panel to date
        (34 models, 10 labs), with a matched core set and explicit data-provenance curation.
  \item We introduce the $h$-field as a per-lab diagnostic that tracks
        release-level capability emphasis and trajectory changes over time.
  \item We show that cooperation cascades: five open-weight families independently
        confirm a second capability transition at 30--72B,
        the isocline phase classifier generalizes across cascade levels,
        and per-lab coupling slopes vary $5\times$.
  \item We provide a three-step diagnostic (locate, classify, predict),
        a per-lab measurement-priority table,
        and seven timestamped falsifiable predictions for the next 12 months.
\end{enumerate}

\noindent\textbf{Terminology.}
The companion paper~\cite{amin2026lying} measures \emph{local coupling}
$\gamma_{12}(N) = \Delta B_2 / \Delta B_1$ between consecutive model sizes within a family---a
derivative that tracks how capabilities interact as models scale.
This paper primarily uses \emph{population coupling} $r(B_1, B_2)$, the Pearson
correlation across a panel of models, which captures cooperative structure across labs.
Both measure the same underlying coupling at different resolutions:
$\gamma_{12}$ detects sign changes within families;
$r$ confirms the cooperative regime persists across the full frontier population.

% ════════════════════════════════════════════════════
\section{Framework: Two Scores, One Diagnostic}
\label{sec:framework}

The diagnostic requires only two benchmark scores per model
and produces three outputs: regime state, lab-relative residual, and transition risk.
All definitions are self-contained; no external reference is required.

\subsection{Capability coupling}

Let $B_1$ and $B_2$ be two benchmark scores measured on the same model.
The \textbf{population coupling} is the Pearson correlation $r(B_1, B_2)$ computed
across a panel of models. When $r > 0$, capabilities reinforce each other
(``cooperative''); when $r < 0$, they trade off (``antagonistic'').

For frontier measurement, we use $B_1 = \text{SWE-bench Verified}$ (autonomous coding)
and $B_2 = \text{GPQA Diamond}$ (graduate-level scientific reasoning).
These axes were chosen because they are widely reported, represent distinct capability
dimensions (code generation vs.\ knowledge-intensive reasoning), and have sufficient
dynamic range at frontier scale.

\subsection{The \texorpdfstring{$h$}{h}-field: lab-relative residual}

Given the population regression $\hat{B}_2 = \beta_1 B_1 + \beta_0$, the \textbf{$h$-field}
for model $i$ is the signed residual:
\begin{equation}
  h_i = B_{2,i} - (\beta_1 B_{1,i} + \beta_0)
  \label{eq:hfield}
\end{equation}
A positive $h_i$ indicates reasoning-rich performance relative to the coding--reasoning trend;
a negative $h_i$ indicates coding-rich. The per-lab mean $\bar{h}_{\rm lab}$ summarizes
a lab's characteristic deviation from the population.

For example, Claude Opus 4.6 has SWE $= 80.8$ and GPQA $= 91.3$.
Using Eq.~\ref{eq:regression}, $h = 91.3 - (0.513 \times 80.8 + 46.4) \approx +3.4$,
placing it slightly reasoning-rich relative to the population trend.

The $h$-field is descriptive, not causal: it summarizes where a model sits relative to
the population trend, not why. It is useful for comparing releases within and across labs
without access to proprietary training details.

\noindent\textbf{Decomposition.}
The aggregate $h$ has internal structure. In the phase-transition framework,
$h$ decomposes into contributions analogous to distinct thermodynamic control parameters:
\begin{equation}
  h_i \;=\; h_{\rm pre} + h_{\rm post} + h_{\rm arch}
  \label{eq:hdecomp}
\end{equation}
where $h_{\rm pre}$ reflects pretraining choices
(data composition, curriculum---permanent changes that shift the phase boundary),
$h_{\rm post}$ reflects post-training interventions
(RLHF, distillation, inference compute---reversible changes
that deflect the system but relax when removed),
and $h_{\rm arch}$ reflects architectural decisions
(width, depth, routing---structural and permanent).
The physics analogy is precise: $h_{\rm pre}$ acts like doping (changes the material),
$h_{\rm post}$ acts like an applied field (changes what is done to the material),
and $h_{\rm arch}$ acts like strain (changes the lattice);
Appendix~\ref{app:physics} develops the correspondence.
At base scale, these are partially separable:
Gemma-4-E4B shows pretraining coupling $0.871$ and RLHF post-training adds
$+0.081$~\cite{amin2026lying}---pretraining contributes ${\sim}10{:}1$
over post-training alignment.
Operationally, $h_{\rm pre}$ shifts persist across releases (they change the model);
$h_{\rm post}$ shifts may reverse (they change what is done to the model).
Section~\ref{sec:results} provides two tests:
the $+7.8$~pp compute-tier shift isolates $h_{\rm post}$,
and Anthropic's three oscillation cycles demonstrate that
post-training excursions relax within one release.

\subsection{Phase classification}

We assign each model to one of three coupling regimes based on $h$:
\begin{itemize}[nosep,leftmargin=*]
  \item \textbf{Coding-rich} ($h < -5$) or \textbf{reasoning-rich} ($h > +5$):
        the model's training emphasis deviates from the population trend.
  \item \textbf{Excursion alert} ($|h| > 10$~pp):
        the deviation is large enough to flag as a capability reallocation
        requiring follow-up (e.g., Sonnet~4.6 at $h = -13.1$).
  \item \textbf{Balanced} ($|h| \leq 5$): near the population trend.
\end{itemize}

\subsection{Isocline phase boundary}
\label{sec:isocline}

The population regression is not merely a statistical fit---it plays the role of
an \emph{isocline phase boundary}.
At base scale, the companion paper~\cite{amin2026lying} shows that the coupling
transition occurs where $\mathrm{TQA}_c = \sqrt{(a/b) \cdot \mathrm{HS}}$,
calibrated from the single independently trained model at $\gamma_{12} = 0$ (OLMo).
The same algebraic structure generalizes to each successive cascade level:
at frontier, $\mathrm{GPQA}_c = \sqrt{0.513 \cdot \mathrm{SWE}}$
(using the regression slope as $a_2/b_2$);
at the next transition, $\mathrm{IFEval}_c = \sqrt{0.97 \cdot \mathrm{GPQA}}$.
Models above the isocline are in the cooperative phase;
models below are entering a new tax regime.
Of the 34 March models, 33 sit above the frontier isocline
(only Claude~3.5 Haiku, the oldest model, falls below).
Section~\ref{sec:cascades} tests the $N_{c,3}$ isocline on models with IFEval scores.

\subsection{Holdout protocol}

We evaluate predictive performance using leave-one-lab-out cross-validation:
for each lab, we fit the regression on all other labs and predict the held-out lab's
GPQA scores from their SWE scores. We report mean absolute error (MAE) across labs.

% ════════════════════════════════════════════════════
\section{Data and Curation Policy}
\label{sec:data}

\subsection{Dataset}

We compile benchmark scores for 34 frontier models from 10 labs
(Anthropic, OpenAI, Google, DeepSeek, Meta, Moonshot, Alibaba, MiniMax, xAI, Zhipu),
spanning releases from June 2024 to March 2026.
We source all scores from official model cards, tech reports, or verified leaderboard
entries (Artificial Analysis, OpenCompass, official blogs).
We do not run evaluations ourselves; this is a measurement paper on public data.

\subsection{Core versus extended subsets}

We maintain a core verified subset ($n = 23$: matched variants, one model per release,
no compute-tier duplicates) and an extended set ($n = 34$: including compute-tier variants,
older anchors, and single-release labs).
Headline claims use the full panel; core serves as a robustness check.

\subsection{Data provenance}

Benchmark scores at the frontier are predominantly self-reported by labs
via model cards, tech reports, and blog posts.
Independent verification lags release by weeks to months.
We flag unverified entries in the extended table and exclude them from core analysis.
The dataset is frozen at a March 2026 cutoff;
models released after this date constitute prospective validation opportunities.

\subsection{Regression specification}

On the full 34-model panel:
\begin{equation}
  \text{GPQA} = (0.513 \pm 0.08) \cdot \text{SWE} + (46.4 \pm 5.2), \quad r = +0.72, \quad p < 10^{-6}
  \label{eq:regression}
\end{equation}
where uncertainties are 95\% confidence intervals on the OLS coefficients.
The residuals have RMSE $= 8.2$~pp, giving a 95\% prediction interval of $\pm 16.2$~pp.
All $h$-field values reported in this paper use Eq.~\ref{eq:regression}.
The regression line also serves as the frontier isocline
(Section~\ref{sec:isocline}): the same algebraic structure
that classifies base models at the coupling boundary~\cite{amin2026lying}.

\begin{figure*}[!htbp]
\centering
\includegraphics[width=\textwidth]{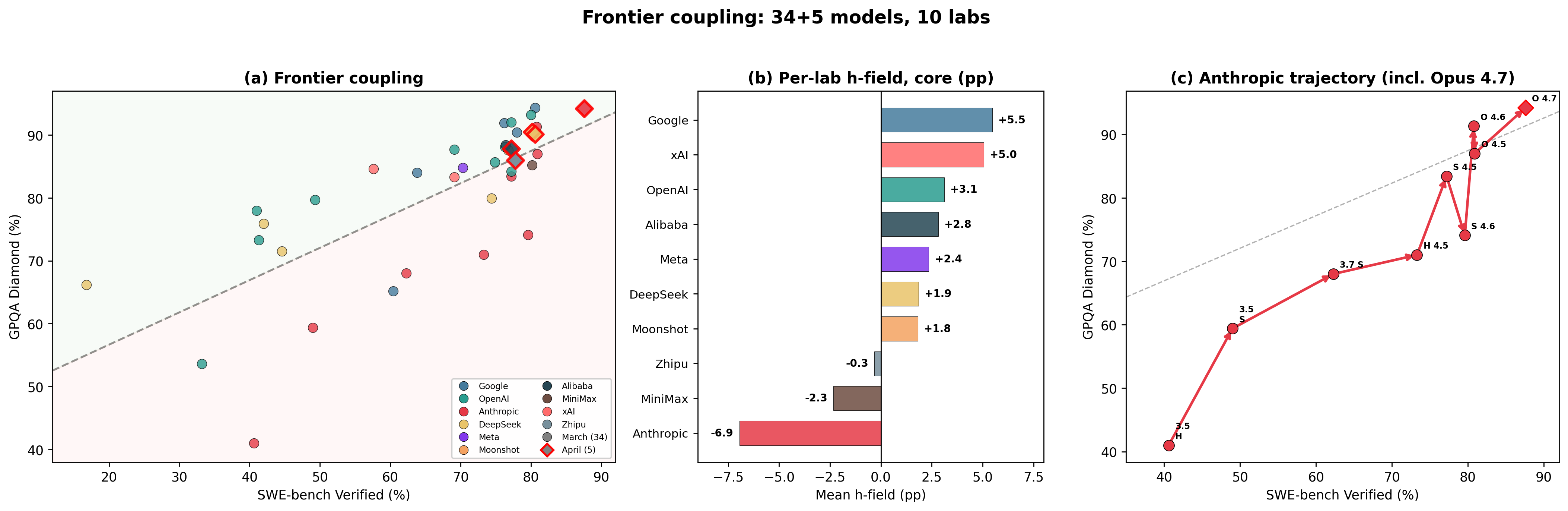}
\caption{\textbf{Frontier coupling: 34 March models + 5 April post-cutoff, 10 labs.}
(a)~SWE-bench Verified vs.\ GPQA Diamond with frozen regression
($\text{GPQA} = 0.513 \cdot \text{SWE} + 46.4$, $r = +0.72$).
Circles: March-frozen models. Diamonds (red edge): April post-cutoff (not used in fit).
(b)~Per-lab $h$-field residual (core models): Google reasoning-rich ($h = +5.5$),
Anthropic coding-rich ($h = -6.9$).
(c)~Anthropic trajectory including post-cutoff Opus~4.7 ($h = +2.9$),
showing three coding--recovery oscillation cycles.}
\label{fig:frontier}
\end{figure*}

% ════════════════════════════════════════════════════
\section{Results}
\label{sec:results}

\subsection{Cooperative coupling at frontier}

SWE-bench Verified and GPQA Diamond are positively coupled across all tested subsets:
\begin{itemize}[nosep,leftmargin=*]
  \item Full panel ($n = 34$): $r = +0.72$, $p < 10^{-6}$
  \item Core verified ($n = 23$): $r = +0.65$, $p < 10^{-3}$
  \item SWE $\geq 40$ ($n = 32$): $r = +0.69$, $p < 10^{-4}$
  \item Excluding compute-tier variants ($n = 32$): $r = +0.72$, $p < 10^{-4}$
\end{itemize}
The cooperative signal is robust to subset definition:
no tested subset produces $r < 0.5$.
That capabilities cooperate at $r > +0.7$ across 10 independent labs---each
with different architectures, training data, distillation pipelines,
and RLHF procedures---is itself the signal:
the cooperative structure persists despite substantial heterogeneity
in how these models were built.
Within each lab, coupling is tighter still ($r > +0.87$ for labs with $\geq 4$ models)
despite smaller size variation than the population---suggesting
the signal reflects recipe-level structure beyond shared scale effects.
All frontier models sit deep in the cooperative regime~\cite{amin2026lying}.

The cooperative structure requires sufficient dynamic range on both benchmark axes
to manifest.
Pre-2025 frontier models span only 16.8--49.0 on SWE-bench ($n = 5$),
compressing the coding axis below the resolution needed to detect coupling.
Models from 2025 onward ($n = 29$, SWE range 42--81) provide
the range in which cooperative structure becomes measurable---consistent
with the base-model finding that coupling emerges only when both
benchmarks are in their informative range.

\subsection{Per-lab \texorpdfstring{$h$}{h}-field diagnostics}

The mean $h$-field per lab reveals systematic release-level capability differences (Fig.~\ref{fig:frontier}):

\begin{table}[h]
\centering
\caption{Per-lab $h$-field and measurement priority.
$n$ = total models available (core + extended + post-cutoff);
$\bar{h}$ is computed on core models only.
Positive $h$ = reasoning-rich; negative = coding-rich.
Confidence: HIGH ($\geq$3 core models, consistent trajectory),
MED (2--3 or recent pivot), LOW (1 model).}
\label{tab:hfield}
\begin{tabular}{lrrllll}
\toprule
Lab & $n$ & $\bar{h}$ & Direction & Trajectory & Next measurement & Conf. \\
\midrule
Google & 5 & $+5.5$ & Reasoning-rich & Consistent & Coding-preserving distill & HIGH \\
OpenAI & 5 & $+3.1$ & Balanced & Ascent & Monitor benchmark rotation & HIGH \\
Alibaba & 1 & $+2.8$ & Balanced & --- & Monitor Qwen3+ trajectory & LOW \\
Meta & 1 & $+2.4$ & Balanced & --- & MoE routing analysis & LOW \\
DeepSeek & 5 & $+1.9$ & Balanced & Oscillation & Track IFEval stabilization & HIGH \\
Moonshot & 1 & $+1.8$ & Balanced & --- & Verify trajectory & LOW \\
Zhipu & 1 & $-0.3$ & Balanced & --- & Collect next release & LOW \\
MiniMax & 1 & $-2.3$ & Balanced & --- & Maintain dual-axis & LOW \\
xAI & 2 & $+5.1$ & Reasoning-rich & --- & Verify next release & LOW \\
Anthropic & 9 & $-6.9$ & Coding-rich & Oscillation & Reasoning preservation & HIGH \\
\bottomrule
\end{tabular}
\end{table}

The lower population correlation ($r = +0.72$) compared to within-lab coupling
($r > +0.87$, Section~\ref{sec:results}) arises because labs sit at different
$h$-field values, not because coupling is weak within any lab.
Per-lab coupling slopes (dGPQA/dSWE) vary $5\times$:
Google ($1.15$) converts each SWE point into $1.15$ GPQA points;
DeepSeek ($0.23$) converts far less efficiently.
This quantifies recipe quality in a single number
(slopes computed from limited release histories, $n = 4$--$10$ per lab;
these are current estimates subject to revision with additional releases).
The $5\times$ variation is consistent with different $h_{\rm pre}$/$h_{\rm post}$ mixtures
(Eq.~\ref{eq:hdecomp}):
Google's high slope reflects efficient distillation from cooperative teachers
(primarily $h_{\rm pre}$);
DeepSeek's low slope suggests architecture-limited transfer
where post-training cannot compensate for weaker pretraining coupling.
This is testable: if a lab improves its slope without retraining
(e.g., by switching inference strategy), the improvement is $h_{\rm post}$;
if it requires a new model generation, it is $h_{\rm pre}$.

The $h$-field also reveals that labs do not simply improve---they pivot,
and the decomposition (Eq.~\ref{eq:hdecomp}) distinguishes transient from structural pivots:

\textbf{DeepSeek reversal} ($h_{\rm pre}$ shift).
DeepSeek's $h$-field drops from $+11.2$ (V2.5) to $-4.7$ (V3.2) across four releases,
a 15.9-pp swing. This monotonic trajectory across model generations
is consistent with an $h_{\rm pre}$ shift---the pretraining pipeline itself
pivoted from reasoning-first to coding-first,
producing a new equilibrium rather than a temporary excursion.

\textbf{Anthropic oscillation} ($h_{\rm post}$ excursions).
Anthropic's trajectory shows an excursion at Claude Sonnet~4.6 ($h = -13.1$),
the deepest coding-specialist deviation in the panel---followed
by recovery at Claude Opus~4.6 ($h = +3.5$),
crossing back to reasoning-rich in a single release cycle.
Three such coding--recovery cycles demonstrate a repeated pattern
consistent with $h_{\rm post}$ dominance: the excursions are release-level
emphasis choices that relax when the emphasis shifts,
while the underlying cooperative coupling persists and re-emerges
at the next generation.

\subsection{Cross-lab holdout}

The framework shows preliminary external validity across labs.
Leave-one-lab-out cross-validation yields $9.2 \pm 2.4$\% MAE (mean $\pm$ SD)
across 4 held-out labs with $\geq 3$ models
(range: OpenAI 6.5\%, Google 7.3\%, DeepSeek 10.6\%, Anthropic 12.4\%).
Prediction is tightest for labs following the population trend and loosest for labs with large
trajectory changes, consistent with the $h$-field capturing real deviations rather than noise.

\subsection{Compute as external field}

Inference compute shifts the $h$-field without retraining,
isolating the $h_{\rm post}$ component (Eq.~\ref{eq:hdecomp}).
GPT-5.4 evaluated at two compute tiers (standard and xhigh):
$\Delta h = +7.8$ pp from standard to xhigh---the same weights, same pretraining,
same architecture, only inference-time compute changes.
This demonstrates that $h_{\rm post}$ alone can move a model
from coding-balanced to reasoning-rich,
and that the diagnostic is sensitive to known interventions.

\subsection{Base-scale bridge}

At base scale, data curation shifts models above the coupling phase boundary:
a 4B distilled model achieves coupling characteristic of 13B+ standard-trained
models~\cite{amin2026lying}.
At frontier scale, per-lab release sequences play the role of model families,
per-lab coupling slopes play the role of per-family trajectories,
and the population regression plays the role of the isocline phase
boundary~\cite{amin2026lying}---the $h$-field measures each lab's deviation
from this boundary, the same physics operating through training recipe
emphasis rather than data curation.

% ════════════════════════════════════════════════════
\section{Capability Cascades at Frontier Scale}
\label{sec:cascades}

\begin{figure*}[!htbp]
\centering
\includegraphics[width=\textwidth]{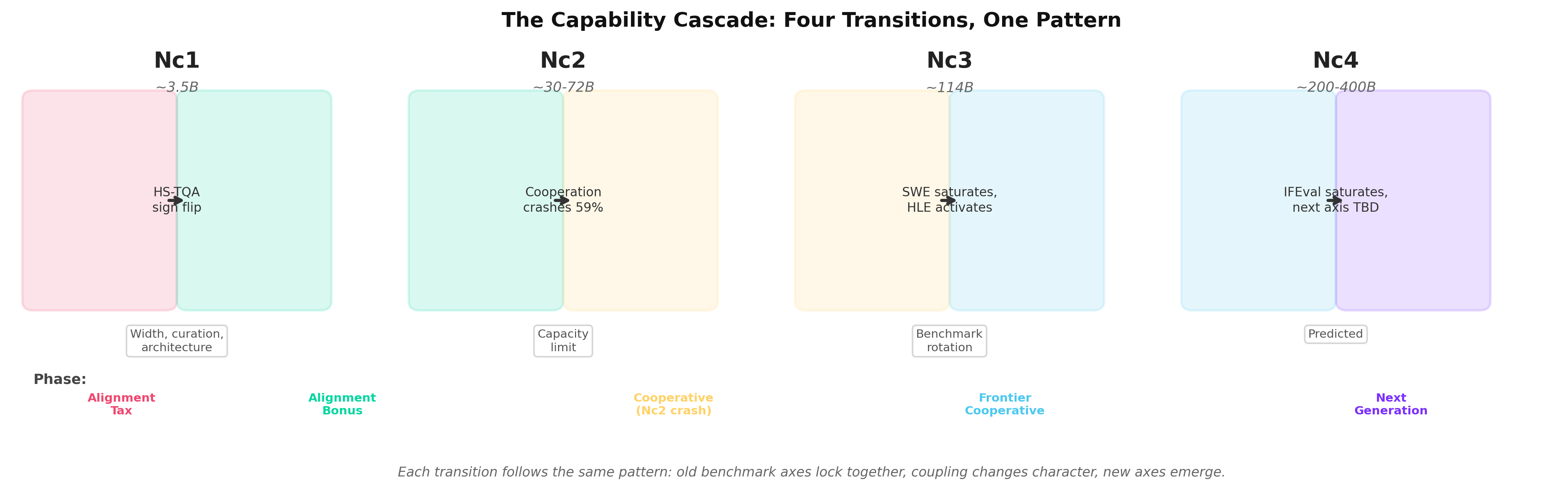}
\caption{\textbf{The capability cascade: four transitions, one pattern.}
At each critical scale, the active benchmark pair changes and coupling undergoes
a qualitative shift. $N_{c,1}$ (${\sim}$3.5B): HS-TQA coupling flips sign.
$N_{c,2}$ (${\sim}$30--72B): cooperation crashes 59\%.
$N_{c,3}$ (${\sim}$114B): SWE saturates, HLE activates.
$N_{c,4}$ (${\sim}$200--400B, predicted): IFEval saturates, next axis TBD.
Engineering levers differ at each transition.}
\label{fig:cascade}
\end{figure*}

The cooperative structure is stable across subsets---but one axis is running out of room.
Cooperation is not static---it cascades through a sequence of transitions
$N_{c,1}, N_{c,2}, N_{c,3}, \ldots$, each following the same pattern:
old benchmark axes lock together, new ones emerge,
and the isocline boundary (Section~\ref{sec:isocline}) identifies the transition
(Fig.~\ref{fig:cascade}).

\subsection{Asymmetric saturation}

The top five coding models are nearly tied: SWE-bench scores span just 1.3 percentage points.
But their reasoning scores span 9.1 points.
When one axis compresses, the population regression pivots to the other,
and previously invisible capability dimensions emerge as new sources of variation.

The multi-benchmark coupling matrix, computed on $n = 9$ common-sample models
with scores on all three benchmarks, confirms this beyond the SWE-GPQA pair:

\begin{table}[h]
\centering
\caption{Frontier coupling matrix on common-sample models ($n = 9$).
All three pairs cooperate, but GPQA--HLE coupling is strongest---GPQA
is the hub connecting coding and expert reasoning.
The matrix is positive definite (eigenvalues: 2.46, 0.42, 0.11).}
\label{tab:coupling}
\begin{tabular}{lrrr}
\toprule
Pair & $r$ & $p$ & $n$ \\
\midrule
SWE--GPQA & $+0.650$ & $0.058$ & 9 \\
GPQA--HLE & $+0.886$ & $0.002$ & 9 \\
SWE--HLE & $+0.649$ & $0.059$ & 9 \\
\bottomrule
\end{tabular}
\end{table}

All three pairs cooperate on the common sample.
GPQA--HLE coupling ($r = +0.886$) is the strongest,
suggesting GPQA is the hub that connects coding capability to expert reasoning.
As SWE-bench scores compress (top-5 spread: 1.3~pp),
GPQA--HLE becomes the more discriminating pair---not because
SWE and HLE decouple, but because GPQA retains the dynamic range
that SWE is losing.
\subsection{$N_{c,2}$: the second transition is measured}

\begin{figure}[!htbp]
\centering
\includegraphics[width=0.95\columnwidth]{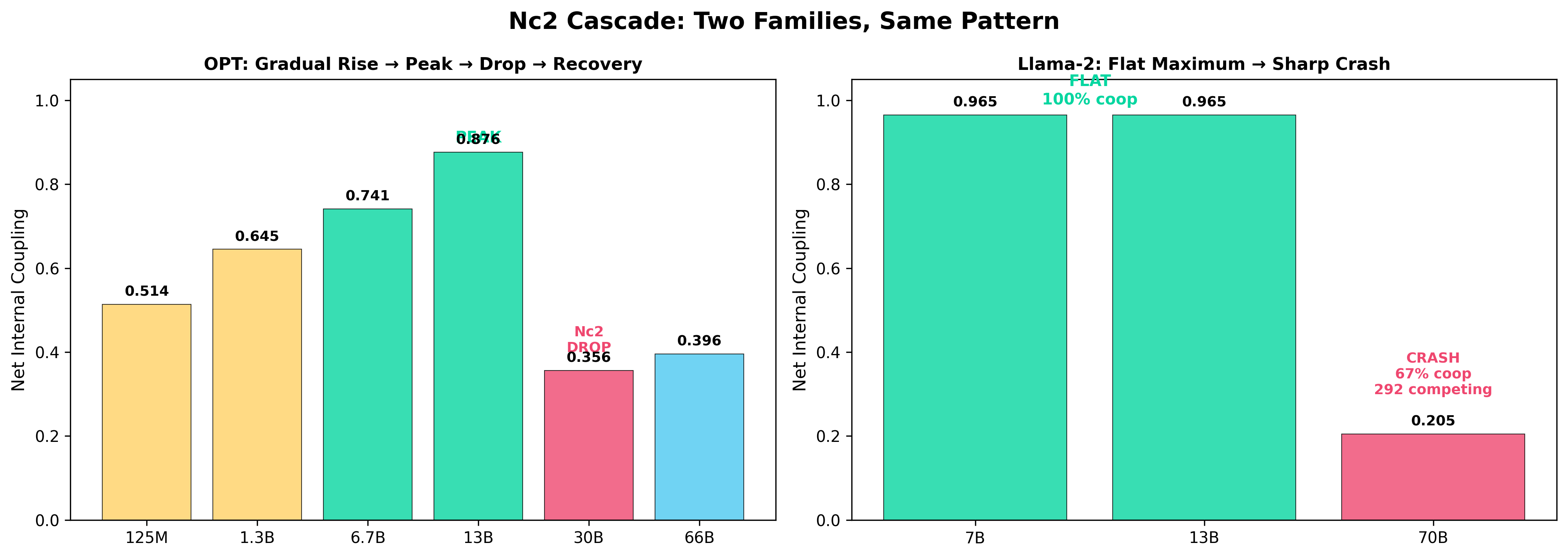}
\caption{\textbf{$N_{c,2}$ cascade: second capability transition at 30--72B.}
OPT (left): gradual rise $\to$ peak at 13B $\to$ drop at 30B $\to$ partial recovery at 66B.
Llama-2 (right): flat maximum at 7B--13B $\to$ sharp crash at 70B.
Same pattern, different mechanism, same net effect.}
\label{fig:nc2}
\end{figure}

The second transition is measured across five open-weight families (Fig.~\ref{fig:nc2}).
OPT's internal cooperation increases monotonically from 125M to 13B
(net coupling $0.514 \to 0.876$, zero competing units),
then drops at 30B ($0.356$, 75 competing units) before partially recovering
at 66B ($0.396$).
Four additional families independently confirm the drop at 30--72B:
Llama-2-70B ($0.205$), Llama-3.1-70B ($0.195$), Qwen2.5-72B ($0.181$),
and OLMo-2-32B ($0.222$).
Three distinct layer-profile patterns---output bottleneck (OPT),
flat weakening (Llama, Qwen), and reversed profile (OLMo-2)---converge
on the same net effect: cooperation weakens when hidden states must encode
more interactions than ${\sim}$30--70B parameters can represent cooperatively.

The character of the transition changes across cascade levels.
At $N_{c,1}$, the coupling transition has near-perfect symmetry
(tax and bonus phases mirror each other~\cite{amin2026lying});
at $N_{c,2}$, the crash is more dramatic---a $59\%$ drop rather than a sign flip---consistent
with an asymmetric transition where the cooperative phase overshoots
its equilibrium before restructuring~\cite{amin2026lying}.
The OPT family spans both transitions in a single ladder:
coupling grows toward a fixed point ($\gamma^* \approx 0.53$),
overshoots to 0.876 at 13B, then crashes to 0.356 at 30B.
The $N_{c,1}$ ODE \emph{fails} at $N_{c,2}$---the fixed point shifts
to $\gamma^*_{N_{c,2}} \approx 0.39$,
and OPT-66B (0.396) is recovering toward it.
Each cascade stage has its own dynamics, not a single extrapolation.

\subsection{$N_{c,3}$: the isocline repeats}

The isocline structure generalizes to the next transition.
At base scale, $\mathrm{TQA}_c = \sqrt{(a/b) \cdot \mathrm{HS}}$
classifies each model as cooperative or tax~\cite{amin2026lying}.
At frontier, $\mathrm{GPQA}_c = \sqrt{0.513 \cdot \mathrm{SWE}}$
does the same (Section~\ref{sec:isocline}).
At the next transition, $\mathrm{IFEval}_c = \sqrt{0.97 \cdot \mathrm{GPQA}}$
is already testable.
Among $n = 4$ frontier models with IFEval scores:
Opus~4.6 sits right at the $N_{c,3}$ boundary (IFEval $94.0\%$ vs.\ boundary $94.1\%$),
Kimi~K2.5 is cooperative ($94.0\% > 92.2\%$),
but MiniMax~M2.5 falls \emph{below} ($87.5\% < 90.9\%$)---consistent
with a coding-specialist training recipe entering a new tax regime
between reasoning and instruction-following.
The mixed-phase regime is already visible:
some models cooperative, one marginal, one below---precisely the distribution
expected at a transition.

The coupling structure (Table~\ref{tab:coupling}) confirms GPQA's role
as the hub connecting current and next axes.
The saturation ratio $\sigma = \text{spread(SWE)} / \text{spread(GPQA)} = 0.14 < 0.2$
signals that the informative channel is already rotating from SWE to the
GPQA--HLE pair---the same dimensional pattern seen at base scale,
where HS$\leftrightarrow$TQA coupling was 0.003 in the tax phase,
0.34 in bonus, and 0.64 at $N_{c,2}$~\cite{amin2026lying}.

The chain of evidence strengthens each prediction:
$N_{c,1}$ was predicted by the coupling framework and confirmed across 16 families;
$N_{c,2}$ was predicted by the cascade hypothesis and observed across 5 open-weight families;
$N_{c,3}$ (SWE saturation, HLE/IFEval activation) now shows preliminary evidence
and is formalized as a falsifiable prediction below.

\section{Predictions}
\label{sec:predictions_section}

\textbf{Per-lab trajectories.}
The $h$-field trajectories (detailed in Section~\ref{sec:results})
reveal that DeepSeek's 15.9-pp reversal and Anthropic's single-release oscillation
are invisible to leaderboard rankings but immediately visible as $h$-field dynamics.
Google's consistent $h \approx +5.5$ across releases suggests a stable
release-level emphasis rather than reactive benchmark optimization.

\begin{figure}[!htbp]
\centering
\includegraphics[width=0.95\columnwidth]{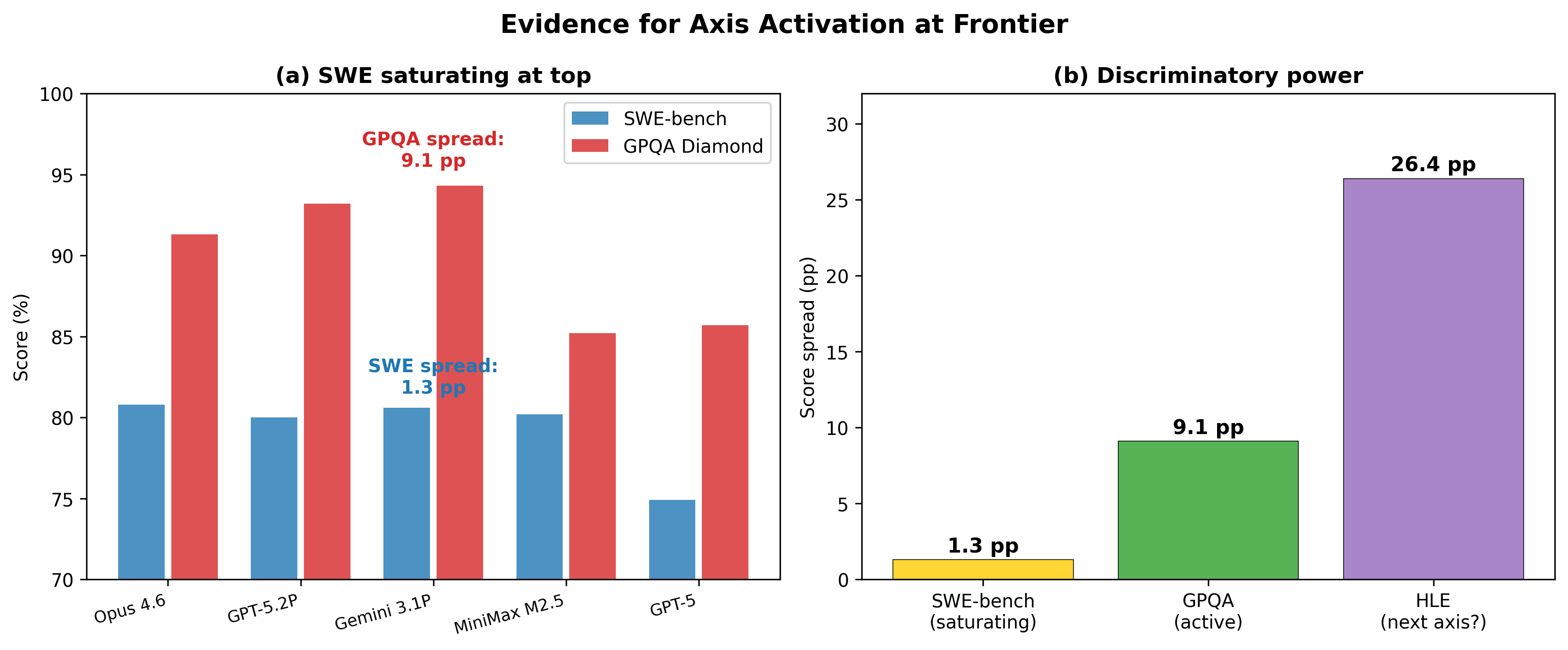}
\caption{\textbf{Asymmetric saturation at the frontier.}
(a)~Among the top-5 SWE-bench models, coding scores compress to a 1.3-pp spread
while GPQA retains 9.1 pp of variation---SWE is losing discriminatory power.
(b)~Benchmark spread among top-5 models: SWE is saturating, GPQA is active,
and HLE (26.4 pp spread) may be the next activating axis.}
\label{fig:nc3}
\end{figure}

\subsection{Seven falsifiable predictions}

We convert each forecast to a timestamped, benchmark-specific test
with pass/fail criteria.
Predictions reference releases after May 2026; post-cutoff models (late March--April 2026)
serve as prospective validation, not prediction targets:

\begin{enumerate}[nosep,leftmargin=*]
  \item \textbf{SWE saturation} (by Dec 2026):
        Top-5 SWE spread $< 2$ pp while GPQA spread $> 5$ pp.
        \textit{Pass}: saturation confirmed. \textit{Fail}: SWE spread $> 5$ pp.

  \item \textbf{IFEval activation} (by Dec 2026):
        $r(\text{GPQA}, \text{IFEval}) > +0.6$ on $n \geq 8$ frontier models.
        \textit{Pass}: new axis confirmed. \textit{Fail}: $r < 0.3$ or $n < 5$.

  \item \textbf{DeepSeek continues coding-first} (next 2 releases):
        $h_{\rm DeepSeek} < 0$ for both next releases.
        \textit{Pass}: trajectory confirmed. \textit{Fail}: $h > +5$ for either.

  \item \textbf{Google maintains reasoning advantage} (next 2 releases):
        $h_{\rm Google} > +3$ for both next releases.
        \textit{Pass}: consistency confirmed. \textit{Fail}: $h < 0$ for either.

  \item \textbf{Cooperative coupling persists} (by May 2027):
        $r(\text{SWE}, \text{GPQA}) > +0.5$ on any frontier panel $\geq 30$ models.
        \textit{Pass}: cooperation is structural. \textit{Fail}: $r < 0.3$.

  \item \textbf{IFEval saturates, HLE activates} ($N_{c,4}$, by Dec 2027):
        IFEval spread among top-10 compresses to $< 3$ pp while HLE spread remains $> 15$ pp.
        \textit{Pass}: $N_{c,4}$ transition underway. \textit{Fail}: IFEval spread $> 8$ pp.

  \item \textbf{GPQA--HLE coupling strengthens relative to SWE--HLE} (by Dec 2026):
        $r(\text{GPQA}, \text{HLE}) > r(\text{SWE}, \text{HLE})$ on $n \geq 10$ frontier models.
        \textit{Pass}: GPQA is the hub connecting to HLE as SWE compresses.
        \textit{Fail}: $r(\text{SWE}, \text{HLE}) > r(\text{GPQA}, \text{HLE})$.
        Current common-sample values: GPQA--HLE $= +0.886$, SWE--HLE $= +0.649$ ($n = 9$).
\end{enumerate}

The predictions form a cascade: SWE saturates (Prediction 1),
IFEval activates alongside it (Prediction 2),
GPQA becomes the dominant hub connecting old and new axes (Prediction 7),
IFEval itself saturates (Prediction 6),
and HLE emerges as the next frontier axis.
Each step follows the same dimensional pattern observed at base-model scale:
old axes lock together, new ones emerge,
and the isocline boundary identifies the transition in real time
(Section~\ref{sec:isocline}).
The framework offers concrete tests for where the frontier
will fracture next.

\textbf{Benchmark rotation: the cascade changes both coupling and coordinates.}
At $N_{c,1}$ (${\sim}$3.5B), the axes stay the same (HS, TQA) but coupling flips sign.
At $N_{c,2}$ (${\sim}$30--72B), coupling crashes within the same axes---a more dramatic
restructuring that changes the fixed point, not just the sign.
Between $N_{c,2}$ and the frontier, the relevant benchmarks themselves shift
from base-model axes (HS, TQA) to frontier axes (SWE, GPQA).
At $N_{c,3}$, the frontier axes rotate again: SWE saturates out (spread/GPQA spread $= 0.14 < 0.2$)
and HLE enters as an independent axis.
The saturation ratio $\sigma = \text{spread}(\text{old}) / \text{spread}(\text{new})$,
where spread is the range (max $-$ min) among the top-5 models on each axis,
acts as a signal-to-noise trigger for benchmark rotation:
$\sigma < 0.2$ signals that the old axis has lost discriminatory power
and the informative channel has rotated to the new pair.
Currently $\sigma(\text{GPQA}/\text{HLE}) = 0.34$---$N_{c,4}$ has not yet begun.

At $N_{c,4}$, we predict IFEval saturates (already at 87--94\%)
and the next discriminating axis will be one of:
AgentBench (multi-step tool use), ARC-AGI-2 (abstract reasoning),
or HarmBench (safety evaluation)---whichever first shows $r > +0.5$
with the currently active axis.
Labs can track this by computing $r(\text{active}, \text{candidate})$
across their release sequence; when $r$ crosses $+0.5$, co-activation has begun.

\textbf{What the isocline predicts.}
At $N_{c,1}$, the coupling transition occurs where $\mathrm{TQA}_c = \sqrt{(a/b) \cdot \mathrm{HS}}$;
at the current frontier, $\mathrm{GPQA}_c = \sqrt{0.513 \cdot \mathrm{SWE}}$
classifies all but one model as cooperative (Section~\ref{sec:isocline}).
The isocline at the next transition, $\mathrm{IFEval}_c = \sqrt{0.97 \cdot \mathrm{GPQA}}$,
already identifies mixed-phase behavior:
two models below the boundary, one at it---suggesting
the next coupling restructuring is underway.
On common-sample models, GPQA--HLE coupling ($r = +0.886$, Table~\ref{tab:coupling})
exceeds SWE--HLE ($r = +0.649$), positioning GPQA as the hub
that will connect the current and next benchmark axes.

\textbf{Practical guidance per model.}
Any practitioner can compute $h$ from two public scores
and determine: (a)~whether their model is coding-heavy or reasoning-heavy
relative to the frontier trend,
(b)~which next measurement or stress test is most informative,
and (c)~whether the current benchmark pair is losing discriminatory power
for their model class.
These three diagnostics require no model internals, no training data access,
and no proprietary information---only two public benchmark scores.
The framework generalizes to any benchmark pair: labs can substitute
internal evaluations, custom safety metrics, or domain-specific benchmarks
for either axis and compute coupling and $h$-field diagnostics identically.

% ════════════════════════════════════════════════════
\section{Related Work}
\label{sec:related}

\textbf{Scaling laws.}
Neural scaling laws predict loss as a power law of compute~\cite{kaplan2020,chinchilla2022}.
These laws are highly precise for aggregate loss but do not address inter-capability interactions.
Observational scaling laws~\cite{ruan2024} extend prediction to individual benchmark scores
from loss proxies, but still treat each benchmark as an independent trajectory.
Our work is complementary: we measure \emph{between}-benchmark coupling, not within-benchmark scaling.

\textbf{Emergent abilities.}
The emergent abilities debate~\cite{wei2022,schaeffer2023} concerns whether capabilities
appear sharply at specific scales or are artifacts of metric choice.
Our framework sidesteps this debate: we measure coupling \emph{between} capabilities,
not emergence \emph{of} individual capabilities.
Whether individual benchmarks show sharp transitions or smooth improvements,
their pairwise coupling can still change regime.

\textbf{Frontier evaluation.}
SWE-bench~\cite{jimenez2024} and GPQA Diamond~\cite{rein2024} have become standard
frontier capability axes.
Chatbot Arena~\cite{chiang2024chatbot} ranks models by human preference but does not
measure inter-capability coupling;
HELM~\cite{liang2023helm} evaluates across many benchmarks but treats each as
an independent trajectory.
Our work complements both: we measure the \emph{coupling between} capability axes,
which is invisible to independent evaluation and irreducible to preference ranking.
Recent work on benchmark saturation~\cite{kiela2021,akhtar2026plateau} and the development of harder
benchmarks (Humanity's Last Exam~\cite{hle2025}) motivates our axis-activation hypothesis:
as one benchmark loses discriminatory power, the informative signal rotates
to the next active axis.
A systematic study of 60 benchmarks~\cite{akhtar2026plateau} finds nearly half exhibit saturation,
with rates increasing as benchmarks age;
our framework provides the coupling mechanism (why saturation cascades)
and rotation protocol (when and what to rotate to).

\textbf{Phase transitions in neural networks.}
Recent theoretical work models phase transitions in linear networks using deformed
Ginzburg-Landau theory with quenched disorder~\cite{arolafernandez2024collective}.
Our approach is empirical rather than theoretical: we measure coupling on real transformer
families and frontier models without requiring a specific theoretical framework.
The base-model foundation is established in~\cite{amin2026lying}.

% ════════════════════════════════════════════════════
\section{Deployment Playbook}
\label{sec:discussion}

The diagnostic reduces to a three-step protocol requiring only
two or three public benchmark scores per model~\cite{amin2026lying}:

\textbf{Step 1: Locate} (any practitioner, 2 minutes).
Compute $h_i$ from Eq.~\ref{eq:hfield} using SWE-bench and GPQA Diamond.
Classify: coding-rich ($h < -5$), balanced ($|h| \leq 5$), or reasoning-rich ($h > +5$).
Flag $|h| > 10$~pp as an excursion alert requiring follow-up.

\textbf{Step 2: Classify} (optional, adds IFEval).
Check whether the model sits above or below the $N_{c,3}$ isocline:
$\mathrm{IFEval} > \sqrt{0.97 \cdot \mathrm{GPQA}}$ (cooperative)
or below (entering a new tax between reasoning and instruction-following).
MiniMax~M2.5 and Qwen3.5 currently fall below; Opus~4.6 sits at the boundary.

\textbf{Step 3: Predict} (requires release history, 10 minutes).
Compare $h_i$ to the lab's previous releases.
Flag shifts $> 10$ pp as capability reallocations requiring follow-up.
\emph{Decompose the shift} (Eq.~\ref{eq:hdecomp}):
if it occurred between model generations (new pretraining), it is $h_{\rm pre}$
and likely persistent---as DeepSeek's four-release reversal demonstrates.
If it occurred within a generation (same base model, different post-training
or compute tier), it is $h_{\rm post}$ and may recover---as Anthropic's
three oscillation cycles confirm.
Check the saturation ratio $\sigma = \text{spread(old)} / \text{spread(new)}$
for the current benchmark pair.

\textbf{Step 3b: Rotate} (organizational decision).
If $\sigma < 0.2$: the current benchmark pair is losing discriminatory power.
Adopt the next benchmark pair---the axis orthogonal to the saturating direction
(currently HLE or IFEval; see Table~\ref{tab:coupling}).
If $|\Delta h| > 10$ pp between releases: treat as a capability reallocation,
verify before changing training policy.

These are diagnostic hypotheses, not causal prescriptions.
A lab optimizing for loss alone may inadvertently train in a regime where
capabilities trade off rather than reinforce~\cite{amin2026lying};
the $h$-field makes this mismatch visible from two public numbers
(Table~\ref{tab:hfield}).
The interactive dashboard at \url{https://zehenlabs.com/cape/}
implements all three levels: enter two benchmark scores to compute $h$ (Level~1),
compare against per-lab release histories (Level~2),
and check saturation ratios for benchmark rotation (Level~3).
It also provides an ODE explorer for per-family trajectory prediction,
a phase classifier for base models, and the isocline boundary test.

\textbf{Worked example.}
Consider Anthropic's Opus~4.7 (SWE $= 87.6$).
The regression predicts GPQA $= 0.513 \times 87.6 + 46.4 = 91.3$.
Actual GPQA $= 94.2$, giving $h = +2.9$: reasoning-rich, recovering from
the Sonnet~4.6 coding excursion ($h = -13.1$).
Had GPQA been $82.0$, then $h = -9.3$: the excursion would have persisted,
and reasoning preservation would be the priority measurement.
The slot-in prediction generalizes:
for any future Anthropic release with SWE $= s$,
the predicted GPQA is $0.513s + 46.4 + \bar{h}_{\rm Anthropic}$,
and the residual from this \emph{lab-specific} baseline flags
whether the release continues the recovery or begins a new excursion.

\textbf{Per-lab outlook (April 2026).}
Five models released after our March data cutoff confirm the diagnostic
out of sample (all within the 95\% prediction interval of $\pm 16.2$~pp;
refitting raises $r$ from $+0.72$ to $+0.75$):
\begin{itemize}[nosep,leftmargin=*]
  \item \textbf{Anthropic} ($\bar{h} = -6.9$, 9 models, HIGH; dominant: $h_{\rm post}$):
    Three coding-specialist excursions (lowest: $h = -13.1$ at Sonnet~4.6)
    with recovery toward reasoning-rich ($h = +3.5$ at Opus~4.6).
    Opus~4.7 ($h = +2.9$) is the third recovery cycle.
    The oscillation pattern is consistent with $h_{\rm post}$ dominance:
    coding emphasis is a release-level choice that relaxes,
    not a pretraining-level shift.
    \emph{Measurement priority}: reasoning-preservation checks after each
    coding-focused release.
  \item \textbf{DeepSeek} ($\bar{h} = +1.9$, 5 models, HIGH; dominant: $h_{\rm pre} \to h_{\rm post}$):
    The initial four-release reversal ($+11.2 \to -4.7$) was $h_{\rm pre}$
    (monotonic across model generations).
    V4~Pro ($h = +2.3$) pivots back, suggesting a new pretraining equilibrium
    with $h_{\rm post}$ oscillation beginning.
    \emph{Measurement priority}: track IFEval to detect whether
    the new equilibrium stabilizes.
  \item \textbf{Google} ($\bar{h} = +5.5$ core, $+5.1$ including extended; 5 models, HIGH; dominant: $h_{\rm pre}$):
    Consistently reasoning-rich across all releases---the signature of stable $h_{\rm pre}$.
    At base scale, Phi achieves cooperative coupling at 1B through data curation~\cite{amin2026lying};
    Google's frontier trajectory mirrors this pattern through distillation,
    sustaining $h > +4$ across every release.
    \emph{Measurement priority}: coding-preserving distillation---whether
    $h > +3$ can be maintained while closing the SWE gap.
  \item \textbf{OpenAI} ($\bar{h} = +3.1$, 10 models, HIGH):
    Steady ascent toward balanced---closest to the population trend.
    GPT-5.4 standard ($h = -1.8$) is the first OpenAI model to dip below the line,
    though the xhigh compute tier recovers to $h = +6.0$ ($\Delta h = +7.8$~pp)---suggesting
    the dip is $h_{\rm post}$ (compute-dependent), not $h_{\rm pre}$.
    \emph{Measurement priority}: as SWE saturates, GPQA--HLE coupling
    becomes the more informative diagnostic.
  \item \textbf{Moonshot} ($\bar{h} = +1.8$ core, $+2.6$ including post-cutoff; 2 models, MED):
    K2.5 $\to$ K2.6: ascending, reasoning-leaning.
    \emph{Measurement priority}: one more release needed to confirm trajectory.
\end{itemize}

\textbf{Base-scale confirmations.}
Three independent confirmations from base-model experiments~\cite{amin2026lying}:
(i)~OLMo (AI2) at $\gamma_{12} = 0.000$ (independent confirmation);
(ii)~Llama-2 cross-prediction at 5.6\% MAE (held-out family);
(iii)~Qwen3 cooperative at all scales (curated training eliminates the tax).
At base scale, targeted activation steering at the CAPE-identified bottleneck
corrects misaligned outputs while leaving already-correct ones unchanged,
with the intervention rate decreasing monotonically from 60\% (tax)
through 30\% (transition) to 20\% (bonus)~\cite{amin2026lying}---the
remaining prompts are true negatives that need no correction,
confirming that intervention efficacy is localized to the predicted regime.

\textbf{From diagnosis to intervention.}
The $h$-field decomposition (Eq.~\ref{eq:hdecomp}) turns
the diagnostic into a prescriptive tool.
The aggregate $h$ tells you \emph{where} a model sits;
the decomposition tells you \emph{what to change}.
If $h_{\rm pre}$ dominates (persistent across releases, as for Google),
improving coupling requires better pretraining---no amount of RLHF
can close a $5\times$ slope gap when pretraining contributes $10{:}1$
over post-training~\cite{amin2026lying}.
If $h_{\rm post}$ dominates (oscillates, as for Anthropic),
the underlying cooperative coupling is intact and will re-emerge;
the priority is monitoring, not retraining.
The GPT-5.4 compute-tier contrast is the cleanest example:
the same weights shift $h$ by $+7.8$~pp between standard and xhigh---pure
$h_{\rm post}$, no retraining, fully reversible.
The per-layer microscopic coupling measured at base scale~\cite{amin2026lying}
(hidden-state coupling $0.725$ compressed to benchmark coupling $0.639$
through the output projection) provides the mechanism:
$h_{\rm pre}$ acts on the hidden states themselves,
while $h_{\rm post}$ acts on how those states are projected to outputs.
Connecting these scales formally---bridging the per-layer ODE to
the macroscopic $h$-field---is the subject of ongoing work.

\textbf{Limitations.}
(1)~The frontier dataset is lab-imbalanced: Anthropic (9 models, core+extended)
and OpenAI (10 models, core+extended)
dominate the panel, while 5 labs contribute a single model each.
(2)~Benchmark scores are predominantly self-reported; independent verification
lags releases.
(3)~The $h$-field captures \emph{what} changed, not \emph{why}---it is descriptive,
not causal. Recommendations in the measurement-priority table
are diagnostic hypotheses, not causal prescriptions.
(4)~$N_{c,3}$ evidence is preliminary ($n = 4$) and classified as such.
(5)~The framework assumes SWE-bench and GPQA Diamond remain meaningful capability axes;
if either benchmark becomes contaminated or saturated, the diagnostic must be
updated with new axes---which the benchmark-rotation protocol is designed to handle.

% ════════════════════════════════════════════════════

\bibliography{references}
\bibliographystyle{plainnat}

% ════════════════════════════════════════════════════
\newpage
\appendix

\section{Methods and Reproducibility}
\label{app:methods}

The frontier dataset is frozen at March 2026 and released as a versioned JSON artifact.
All regression equations, $h$-field calculations, and holdout protocols
are specified in Sections~\ref{sec:framework}--\ref{sec:data} with sufficient detail
for independent reproduction.
Prediction pass/fail criteria (Section~\ref{sec:predictions_section}) are timestamped
and benchmark-specific to enable unambiguous future evaluation.

Population coupling $r(B_1, B_2)$ is the Pearson correlation across the model panel.
The $h$-field (Eq.~\ref{eq:hfield}) is the signed residual from OLS regression
$\text{GPQA} = 0.513 \cdot \text{SWE} + 46.4$ ($r = +0.72$, $n = 34$, $p < 10^{-6}$),
frozen at the March 2026 cutoff.
The decomposition (Eq.~\ref{eq:hdecomp}) has three evidence levels:
$h_{\rm post}$ is directly isolated by the compute-tier experiment
($\Delta h = +7.8$~pp, same weights);
$h_{\rm pre}$ is inferred from cross-generation comparisons
(Gemma-4: base $0.871$, instruct $0.952$~\cite{amin2026lying})
and per-lab trajectory patterns;
$h_{\rm arch}$ is hypothesized (no isolation experiment in this study).
Leave-one-lab-out cross-validation yields $9.2 \pm 2.4$\% MAE
across 4 labs with $\geq 3$ models.
Saturation ratio $\sigma = \text{spread(old)} / \text{spread(new)}$
computed on top-5 models per benchmark.
The isocline classifier $B_{2,c} = \sqrt{(a/b) \cdot B_1}$ uses
$a/b = 0.513$ (the regression slope) at the SWE--GPQA transition
and $a_3/b_3 = 0.97$ at the GPQA--IFEval transition~\cite{amin2026lying}.
The multi-benchmark coupling matrix (Table~\ref{tab:coupling}) is computed
on $n = 9$ common-sample models with all three scores (SWE, GPQA, HLE);
the matrix is verified positive definite (eigenvalues: 2.46, 0.42, 0.11).
Residual RMSE $= 8.2$~pp; 95\% prediction interval $= \pm 16.2$~pp.

\section{Full Frontier Model Table}
\label{app:table}

\begin{table}[h]
\centering
\caption{Complete frontier panel (34 March + 5 April post-cutoff, 10 labs). Core models marked with $\star$.
$h$-field computed from frozen 34-model regression: $\text{GPQA} = 0.513 \cdot \text{SWE} + 46.4$.}
\label{tab:full}
\begin{tabular}{llrrrr}
\toprule
Model & Lab & SWE & GPQA & $h$ & Subset \\
\midrule
Claude 3.5 Haiku & Anthropic & 40.6 & 41.0 & $-26.2$ & Extended \\
Claude 3.5 Sonnet & Anthropic & 49.0 & 59.4 & $-12.1$ & $\star$ Core \\
Claude 3.7 Sonnet & Anthropic & 62.3 & 68.0 & $-10.3$ & $\star$ Core \\
Claude Haiku 4.5 & Anthropic & 73.3 & 71.0 & $-13.0$ & $\star$ Core \\
Claude Sonnet 4.5 & Anthropic & 77.2 & 83.4 & $-2.6$ & $\star$ Core \\
Claude Opus 4.5 & Anthropic & 80.9 & 87.0 & $-0.9$ & $\star$ Core \\
Claude Sonnet 4.6 & Anthropic & 79.6 & 74.1 & $-13.1$ & $\star$ Core \\
Claude Opus 4.6 & Anthropic & 80.8 & 91.3 & $+3.5$ & $\star$ Core \\
DeepSeek-V2.5 & DeepSeek & 16.8 & 66.2 & $+11.2$ & Extended \\
DeepSeek-V3 & DeepSeek & 42.0 & 75.9 & $+8.0$ & $\star$ Core \\
DeepSeek-R1 & DeepSeek & 44.6 & 71.5 & $+2.2$ & $\star$ Core \\
DeepSeek V3.2 & DeepSeek & 74.4 & 79.9 & $-4.7$ & $\star$ Core \\
Gemini 2.0 Flash & Google & 60.4 & 65.2 & $-12.1$ & Extended \\
Gemini 2.5 Pro & Google & 63.8 & 84.0 & $+4.9$ & $\star$ Core \\
Gemini 3 Flash & Google & 78.0 & 90.4 & $+4.0$ & $\star$ Core \\
Gemini 3 Pro & Google & 76.2 & 91.9 & $+6.4$ & $\star$ Core \\
Gemini 3.1 Pro & Google & 80.6 & 94.3 & $+6.6$ & $\star$ Core \\
Llama 4 Maverick & Meta & 70.3 & 84.8 & $+2.4$ & $\star$ Core \\
MiniMax M2.5 & MiniMax & 80.2 & 85.2 & $-2.3$ & $\star$ Core \\
Kimi K2.5 & Moonshot & 76.8 & 87.6 & $+1.8$ & $\star$ Core \\
Qwen3.5-397B & Alibaba & 76.4 & 88.4 & $+2.8$ & $\star$ Core \\
GPT-4o & OpenAI & 33.2 & 53.6 & $-9.8$ & Extended \\
o1-preview & OpenAI & 41.3 & 73.3 & $+5.7$ & Extended \\
o1 & OpenAI & 41.0 & 78.0 & $+10.6$ & Extended \\
o3-mini & OpenAI & 49.3 & 79.7 & $+8.0$ & $\star$ Core \\
o3 & OpenAI & 69.1 & 87.7 & $+5.8$ & $\star$ Core \\
GPT-5 & OpenAI & 74.9 & 85.7 & $+0.9$ & $\star$ Core \\
GPT-5.1 & OpenAI & 76.3 & 88.1 & $+2.6$ & $\star$ Core \\
GPT-5.2 Pro & OpenAI & 80.0 & 93.2 & $+5.9$ & Extended \\
GPT-5.4 std & OpenAI & 77.2 & 84.2 & $-1.8$ & $\star$ Core \\
GPT-5.4 xhigh & OpenAI & 77.2 & 92.0 & $+6.0$ & Extended \\
Grok 3 & xAI & 57.6 & 84.6 & $+8.7$ & Extended \\
Grok 4 & xAI & 69.1 & 83.3 & $+1.5$ & Extended \\
GLM-5 & Zhipu & 77.8 & 86.0 & $-0.3$ & Extended \\
\midrule
\multicolumn{6}{l}{\emph{Post-cutoff (April 2026, not used in frozen regression)}} \\
Claude Opus 4.7 & Anthropic & 87.6 & 94.2 & $+2.9$ & Post-cutoff \\
Kimi K2.6 & Moonshot & 80.2 & 90.5 & $+2.9$ & Post-cutoff \\
DeepSeek V4 Pro & DeepSeek & 80.6 & 90.1 & $+2.3$ & Post-cutoff \\
Qwen3.6-27B & Alibaba & 77.2 & 87.8 & $+1.8$ & Post-cutoff \\
GLM-5.1 & Zhipu & 77.8 & 86.0 & $-0.3$ & Post-cutoff \\
\bottomrule
\end{tabular}
\end{table}

\section{Base-Model Context}
\label{app:base}

\begin{figure}[h]
\centering
\includegraphics[width=0.95\columnwidth]{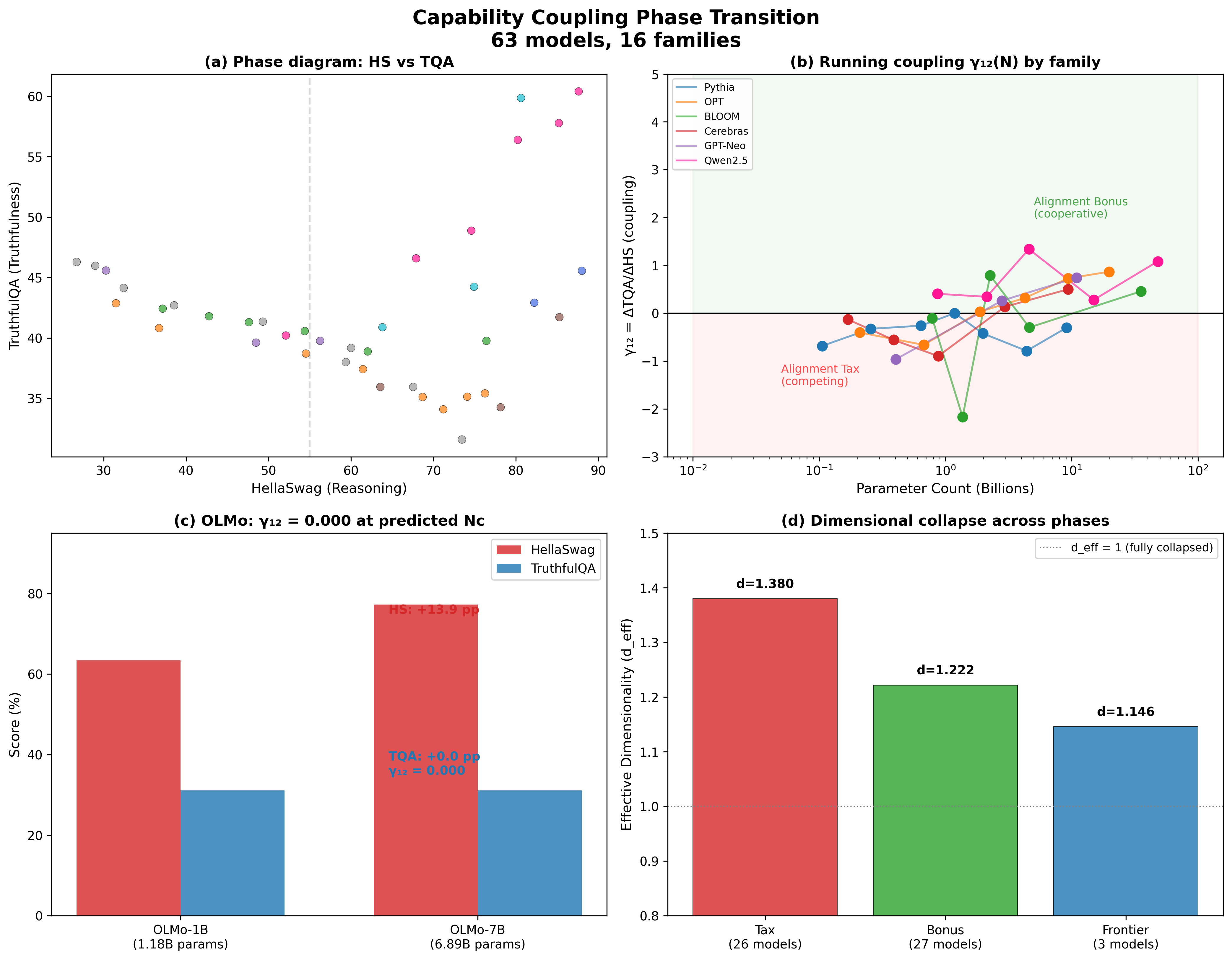}
\caption{\textbf{Base-model foundation (from~\cite{amin2026lying}).}
The coupling regime transition underlying CAPE: below a critical scale,
reasoning and truthfulness anticorrelate; above, they cooperate.
All frontier models sit in the cooperative regime.}
\label{fig:base}
\end{figure}

\section{Extended Robustness}
\label{app:robustness}

\textbf{Sensitivity to subset definition.}
The cooperative signal is robust across subset choices:
$r = +0.72$ (full 34), $r = +0.65$ (core 23), $r = +0.69$ (SWE $\geq 40$),
$r = +0.72$ (excluding compute-tier variants).
No subset produces $r < 0.5$, confirming that cooperative coupling is robust to
inclusion criteria rather than an artifact of specific model selection.

\textbf{Dynamic range dependence.}
The cooperative structure requires sufficient dynamic range on both benchmark axes.
Pre-2025 frontier models span only 16.8--49.0 on SWE-bench ($n = 5$),
compressing the coding axis below the resolution at which coupling becomes measurable.
Models from 2025 onward ($n = 29$, SWE range 42--81) provide the range
in which cooperative structure manifests.

\section{Physics Analogy (Optional Context)}
\label{app:physics}

For readers familiar with condensed-matter physics, the CAPE framework has formal
parallels to Ginzburg-Landau theory of multi-band superconductors:
benchmark scores play the role of order parameters,
$\log_{10} N$ plays the role of temperature,
and the $h$-field plays the role of an external magnetic field breaking phase symmetry.
The coupling $\gamma_{12}$ corresponds to inter-band pairing susceptibility.
At the base-model level, this analogy is quantitatively precise
(12 diagnostics from 2 parameters; see~\cite{amin2026lying}).
At the frontier, we use only the operational definitions from Section~\ref{sec:framework};
the physics analogy provides interpretive context but is not required for any claim.

\section*{Broader Impact}
This work provides a public diagnostic for frontier model development.
The $h$-field allows outside observers to track whether a lab's releases are becoming
more coding-specialist or more reasoning-balanced without access to training details.
This transparency could inform policy discussions about frontier model trajectories.
Two public numbers per model are enough to start seeing where the frontier is going.

\end{document}